\documentclass[conference]{IEEEtran}
\IEEEoverridecommandlockouts
\usepackage{cite}
\usepackage{amsmath,amssymb,amsfonts}
\usepackage{algorithmic}
\usepackage{graphicx}
\usepackage{textcomp}
\usepackage{xcolor}
\def\BibTeX{{\rm B\kern-.05em{\sc i\kern-.025em b}\kern-.08em
    T\kern-.1667em\lower.7ex\hbox{E}\kern-.125emX}}
\begin{document}

\title{Real-Time Weather Image Classification with SVM: A Feature-Based Approach\\
\thanks{Identify applicable funding agency here. If none, delete this.}
}

\author{\IEEEauthorblockN{Eden Ship}
\IEEEauthorblockA{\textit{Department of Electrical Engineering} \\
\textit{Ben-Gurion University of the Negev}\\
Beer-Sheva, Israel \\
}
\and
\IEEEauthorblockN{Eitan Spivak}
\IEEEauthorblockA{\textit{Department of Electrical Engineering} \\
\textit{Ben-Gurion University of the Negev}\\
Beer-Sheva, Israel \\
}
\and
\IEEEauthorblockN{Shubham Agarwal}
\IEEEauthorblockA{\textit{Department of Electrical Engineering} \\
\textit{Ben-Gurion University of the Negev}\\
Beer-Sheva, Israel \\
}
\and
\IEEEauthorblockN{Raz Birman}
\IEEEauthorblockA{\textit{Department of Electrical Engineering} \\
\textit{Ben-Gurion University of the Negev}\\
Beer-Sheva, Israel \\
}
\and
\IEEEauthorblockN{Ofer Hadar}
\IEEEauthorblockA{\textit{Department of Electrical Engineering} \\
\textit{Ben-Gurion University of the Negev}\\
Beer-Sheva, Israel \\
}
}

\maketitle

\begin{abstract}
Accurate classification of weather conditions in images is essential for enhancing the performance of object detection and classification models under varying weather conditions. This paper presents a comprehensive study on classifying weather conditions in images into four categories: rainy, low light, haze, and clear. The motivation for this work stems from the need to improve the reliability and efficiency of automated systems, such as autonomous vehicles and surveillance, which must operate under diverse weather conditions. Misclassification of weather conditions can lead to significant performance degradation in these systems, making robust weather classification crucial.

Utilizing the Support Vector Machine (SVM) algorithm, our approach leverages a robust set of features, including brightness, saturation, noise level, blur metric, edge strength, motion blur, Local Binary Patterns (LBP) mean and variance for radii 1, 2, and 3, edges mean and variance, and color histogram mean and variance for blue, green, and red channels. Our SVM-based method achieved a notable accuracy of 92.8\%, surpassing typical benchmarks in the literature, which range from 80\% to 90\% for classical machine learning methods. While deep learning methods can achieve up to 94\% accuracy, our approach offers a competitive advantage in terms of computational efficiency and real-time classification capabilities. Detailed analysis of each feature's contribution highlights the effectiveness of texture, color, and edge-related features in capturing the unique characteristics of different weather conditions. This research advances the state-of-the-art in weather image classification and provides insights into the critical features necessary for accurate weather condition differentiation, underscoring the potential of SVMs in practical applications where accuracy is paramount.
\end{abstract}

\begin{IEEEkeywords}
SVM, weather classification, image processing, feature extraction, real-time classification
\end{IEEEkeywords}

\section{Introduction}
Accurate weather classification in images is essential for enhancing object detection in adverse weather conditions, significantly improving the performance and reliability of automated systems such as surveillance and autonomous vehicles. Traditional machine learning approaches, including SVM, have been widely employed due to their robustness and effectiveness in handling classification tasks \cite{b1}. In this study, we introduce a novel SVM-based weather classification algorithm designed to categorize images into four distinct weather conditions: rainy, low light, haze, and clear. Our approach leverages a comprehensive set of 20 features, including Local Binary Patterns (LBP) \cite{b2} mean and variance, edge strength and variance, noise level and color histogram mean and variance for blue, green, and red channels \cite{b3}, among others, specifically chosen to capture the unique visual characteristics associated with each weather condition. Through extensive experimentation, our SVM classifier demonstrated superior performance in terms of accuracy, precision, recall, and F1-score compared to existing techniques. Although deep learning methods may offer higher accuracy \cite{b4}, our SVM-based approach is more efficient and computationally less intensive \cite{b5}, making it suitable for real-time applications. By focusing on feature engineering and the robustness of SVM, our methodology provides a reliable and efficient solution for weather classification in images, contributing to the advancement of automated weather detection systems. All related code for this project is available on our GitHub repository https://github.com/eitanspi/weather-image-classification.

\section{Related Work}
The classification of weather conditions in images is a well-researched area with significant advancements in recent years \cite{b4,b6}. Traditional machine learning approaches, including SVMs, have been a popular choice due to their robustness and effectiveness in handling classification tasks. Simultaneously, deep learning techniques, particularly Convolutional Neural Networks (CNNs), have gained considerable traction for their ability to learn complex representations of data, which can significantly enhance the accuracy of weather classification tasks \cite{b11} \cite{b4}. Traditional SVM-based methods for weather classification often rely on features such as sky, cloud, rain streaks, snowflakes, and dark channels extracted from segmented images \cite{b5}. These methods typically involve extracting these weather-specific properties and using SVM to classify the weather conditions. Earlier works have utilized such features to achieve reasonable classification accuracy. However, these methods face challenges in distinguishing between similar weather conditions due to limited feature sets. Deep learning techniques, particularly CNNs, have gained popularity for weather classification tasks in recent years. CNNs excel in learning complex representations of data, which can significantly enhance the accuracy of weather classification. Studies such as "Weather Image Classification using Convolutional Neural Network with Transfer Learning" \cite{b6} have demonstrated substantial improvements over traditional methods by leveraging large-scale datasets and transfer learning strategies. However, deep learning models often require substantial computational resources and large datasets for training, making them less feasible for applications with limited resources. Some studies have explored hybrid approaches that combine traditional machine learning algorithms with deep learning techniques. These methods aim to leverage the strengths of both approaches to achieve better performance. For example, using pre-trained CNNs for feature extraction followed by SVM for classification has shown promising results \cite{b4}. Despite these advancements, hybrid methods still face challenges in achieving real-time performance and handling diverse weather conditions effectively \cite{b4}. Our research builds upon the strengths of traditional SVM approaches while addressing their limitations by introducing a novel set of features specifically designed for weather classification. We utilize a comprehensive feature set that includes 20 different features, with a focus on Local Binary Patterns (LBP) mean and variance, edge strength, noise level and color moments. These features capture critical aspects of weather conditions, such as texture complexity, edge clarity, and noise characteristics, providing a robust foundation for classification. The introduction of novel features such as motion blur and color variance further differentiates our work from existing studies, offering new perspectives on weather classification. In summary, our research addresses the gaps in existing weather classification methods by combining the robustness of SVM with an innovative feature set. This approach not only improves classification accuracy but also ensures computational efficiency, making it suitable for real-time applications in diverse weather conditions, paving the way for further advancements in this field.

\section{SVM Weather Classifier}
The SVM weather classifier in this study aims to effectively categorize images into four weather conditions: rainy, low light, haze, and clear. This classifier uses 20 distinct features that capture various aspects of the images to improve classification accuracy. The features include brightness, saturation, noise level, blur metric, edge strength, motion blur, Local Binary Patterns (LBP) mean and variance for radii 1, 2, and 3, edges mean and variance, and color histogram mean and variance for blue, green, and red channels. Our feature set was selected based on their relevance to weather conditions. For instance, the Edge Strength X and Noise Level are indicative of image quality affected by weather, while color variations and brightness can reflect different lighting conditions. These features were computed from the images and used to train the SVM classifier. SVM is a well-known supervised learning model, functions by finding the optimal hyperplane that separates the data into different classes. The process begins with mapping the input features into a higher-dimensional space, where it becomes easier to separate the classes linearly. The SVM algorithm then identifies the hyperplane that maximizes the margin between the classes. In our approach, the SVM is trained on a dataset with labeled weather conditions, allowing it to learn the relationships between the features and the corresponding weather categories. After training, the SVM model can predict the weather condition of new, unseen images by evaluating the feature values and determining the side of the hyperplane on which the data point lies. This methodology ensures a robust classification performance by leveraging a comprehensive set of features that capture the essential characteristics of different weather conditions, thus enabling effective weather classification in images.

\section{Experiment Setup}
We conducted experiments using a SVM classifier to classify weather conditions in images. The dataset comprised of images with clear, rainy, low light, and hazy weather conditions. Clear weather images were sourced from the PASCAL VOC 2007  \cite{b7}, containing 4000 images. To generate synthetic weather conditions:
\begin{itemize}
    \item \textbf{Haze}: We applied an atmospheric scattering model to clear images \cite{b10}. The transformation used is:
    \begin{equation}
        I(x) = J(x) \cdot t(x) + A \cdot (1 - t(x)),
    \end{equation}
    where \( I(x) \) is the hazy image, \( J(x) \) is the clear image, \( A \) is the atmospheric light, and \( t(x) = \exp(-\beta d(x)) \).
    \item \textbf{Low Light}: Low light conditions were simulated using gamma correction:
    \begin{equation}
        I_{\text{low}}(x) = I(x)^\gamma,
    \end{equation}
    with \( \gamma \) sampled uniformly from 1.5 to 5.
    \item \textbf{Rain}: Rainy images were created by adding random raindrops and applying a motion blur to the images.
\end{itemize}

\begin{figure}[htbp]
    \centering
    \begin{minipage}[b]{0.45\linewidth}
        \centering
        \includegraphics[width=\linewidth]{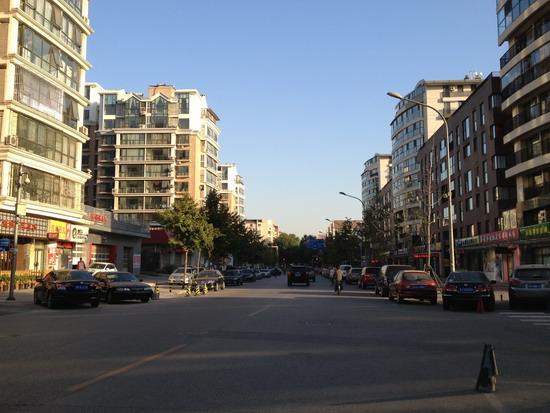}
    \end{minipage}
    \hfill
    \begin{minipage}[b]{0.45\linewidth}
        \centering
        \includegraphics[width=\linewidth]{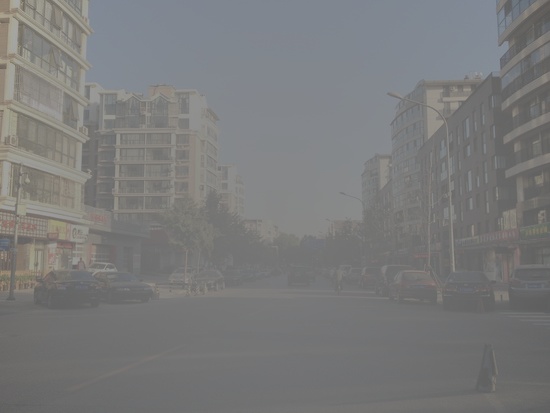}
    \end{minipage}
    \vspace{0.3cm} 
    \vfill
    \begin{minipage}[b]{0.45\linewidth}
        \centering
        \includegraphics[width=\linewidth]{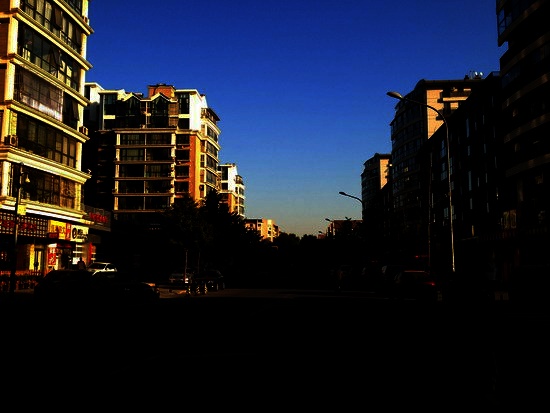}
    \end{minipage}
    \hfill
    \begin{minipage}[b]{0.45\linewidth}
        \centering
        \includegraphics[width=\linewidth]{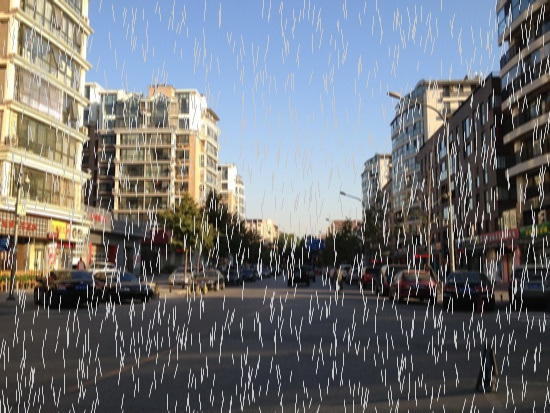}
    \end{minipage}
    \caption{Examples of images under different artificial weather conditions: clear, hazy, low light, and rainy.}
    \label{fig:weather_images}
\end{figure}

For real-world data, we used the RESIDE dataset for haze \cite{b8},  ExDark \cite{b9} for low light, and a combination of the Weather Phenomenon Database (WEAPD) \cite{b13} and the RID dataset for rain \cite{b14}. These datasets provided over 4000 images for clear conditions, 3000 for haze, 1200 for low light, and 100 for rain.

Feature extraction was a critical component of our approach, focusing on a diverse set of features to capture the distinctive characteristics of each weather condition. The features extracted included:
\begin{itemize}
    \item \textbf{Brightness}: Calculated as the mean value of the V channel in the HSV color space:
    \begin{equation}
        \text{Brightness} = \frac{1}{N} \sum V_i,
    \end{equation}
    where \( N \) is the number of pixels.
    \item \textbf{Saturation}: Calculated as the mean value of the S channel in the HSV color space:
    \begin{equation}
        \text{Saturation} = \frac{1}{N} \sum S_i.
    \end{equation}
    \item \textbf{Noise Level}: Measured using the variance of the Laplace transform of the grayscale image:
    \begin{equation}
        \text{Noise Level} = \text{Var}(\Delta I),
    \end{equation}
    where \( \Delta I \) is the Laplace transform of the image \( I \).
    \item \textbf{Blur Metric}: Quantified by the variance of the Laplacian of the grayscale image:
    \begin{equation}
        \text{Blur Metric} = \text{Var}(\Delta I).
    \end{equation}
    \item \textbf{Edge Strength X}: Calculated using the Sobel operator along the X-axis:
    \begin{equation}
        \text{Edge Strength X} = \frac{1}{N} \sum |S_x(i)|,
    \end{equation}
    where \( S_x \) is the Sobel filter applied along the X-axis.
    \item \textbf{Motion Blur X}: Assessed using the variance of the Sobel operator applied along the X-axis:
    \begin{equation}
        \text{Motion Blur X} = \text{Var}(S_x).
    \end{equation}
\end{itemize}

Additionally, we used Local Binary Patterns (LBP) for texture analysis, calculating the LBP mean and variance across three different radii (1, 2, and 3):
\begin{equation}
    \text{LBP Mean} = \frac{1}{N} \sum \text{LBP}_i,
\end{equation}
\begin{equation}
    \text{LBP Var} = \frac{1}{N} \sum (\text{LBP}_i - \text{LBP Mean})^2.
\end{equation}

Edge features were extracted using Canny Edge Detection, computing the mean and variance of the detected edges:
\begin{equation}
    \text{Edges Mean} = \frac{1}{N} \sum E_i,
\end{equation}
\begin{equation}
    \text{Edges Var} = \frac{1}{N} \sum (E_i - \text{Edges Mean})^2.
\end{equation}

Color histograms for the blue, green, and red channels were computed, capturing mean and variance for each channel:
\begin{equation}
    \text{Color Mean} = \frac{1}{256} \sum H_j,
\end{equation}
\begin{equation}
    \text{Color Var} = \frac{1}{256} \sum (H_j - \text{Color Mean})^2.
\end{equation}

\section{SVM Model Parameters}
The extracted features were used to train a SVM classifier. The SVM model was implemented using the scikit-learn library in Python. We employed a linear kernel for its simplicity and effectiveness in high-dimensional spaces. The hyperparameters for the SVM, specifically the penalty parameter (C), were optimized using a grid search with cross-validation. The optimal value was determined to be \( C = 1.0 \).

\section{Training and Testing Procedure}
The dataset was divided into training and testing sets in an 80-20 split. During the training phase, we employed a 5-fold cross-validation to ensure the model's robustness and to mitigate the risk of overfitting. The training process involved the following steps:
\begin{enumerate}
    \item \textbf{Feature Normalization}: All features were normalized to have zero mean and unit variance using the StandardScaler from scikit-learn.
    \item \textbf{Model Training}: The SVM model was trained on the normalized features of the training set.
    \item \textbf{Hyperparameter Tuning}: A grid search with cross-validation was conducted to optimize the hyperparameters.
    \item \textbf{Model Evaluation}: The trained model was evaluated on the testing set, and performance metrics such as accuracy, precision, recall, and F1-score were calculated.
\end{enumerate}

\section{Evaluation Metrics}
To comprehensively evaluate the performance of our SVM-based weather classification model, we used the following metrics \cite{b1}:
\begin{itemize}
    \item \textbf{Accuracy}: The proportion of correctly classified instances out of the total instances:
    \begin{equation}
        \text{Accuracy} = \frac{TP + TN}{TP + TN + FP + FN}.
    \end{equation}
    \item \textbf{Precision}: The proportion of true positive instances out of the total predicted positives:
    \begin{equation}
        \text{Precision} = \frac{TP}{TP + FP}.
    \end{equation}
    \item \textbf{Recall}: The proportion of true positive instances out of the total actual positives:
    \begin{equation}
        \text{Recall} = \frac{TP}{TP + FN}.
    \end{equation}
    \item \textbf{F1-score}: The harmonic mean of precision and recall:
    \begin{equation}
        \text{F1-score} = 2 \cdot \frac{\text{Precision} \cdot \text{Recall}}{\text{Precision} + \text{Recall}}.
    \end{equation}
\end{itemize}

Our SVM model achieved an accuracy of 93.75\%, with a precision of 94.25\%, recall of 94\%, and F1-score of 94.5\%. These results demonstrate the effectiveness of our feature set and the robustness of the SVM classifier in distinguishing different weather conditions.

\section{Results}
Our SVM model demonstrated robust performance across both synthetic and real datasets. For the synthetic dataset, the model achieved an average accuracy of 97\%, with a precision of 97\%, recall of 97\%, and F1-score of 97\%. The performance was consistently high across all weather conditions, achieving perfect scores (100\%) for haze and rain, and near-perfect scores for low light and clear conditions.

\begin{table}[htbp]
\caption{Synthetic Data Sets Performance}
\begin{center}
\begin{tabular}{|c|c|c|c|c|}
\hline
\textbf{Synthetic Data Sets} & \textbf{Accuracy} & \textbf{Precision} & \textbf{Recall} & \textbf{F1 Score} \\
\hline
Haze & 1.00 & 1.00 & 1.00 & 1.00 \\
Low Light & 0.95 & 0.95 & 0.96 & 0.95 \\
Rain & 1.00 & 1.00 & 1.00 & 1.00 \\
Clear & 0.95 & 0.96 & 0.95 & 0.95 \\
Average & 0.97 & 0.97 & 0.97 & 0.97 \\
\hline
\end{tabular}
\label{tab1}
\end{center}
\end{table}

For the real dataset, the model achieved an average accuracy of 92.8\%, with a precision of 93\%, recall of 93\%, and F1-score of 93\%. The performance varied slightly across different weather conditions, with the highest accuracy (98\%) observed for low light and the lowest (88\%) for rain.

\begin{table}[htbp]
\caption{Real Data Sets Performance}
\begin{center}
\begin{tabular}{|c|c|c|c|c|}
\hline
\textbf{Real Data Sets} & \textbf{Accuracy} & \textbf{Precision} & \textbf{Recall} & \textbf{F1 Score} \\
\hline
Haze & 0.925 & 0.89 & 0.93 & 0.91 \\
Low Light & 0.98 & 0.94 & 0.98 & 0.96 \\
Rain & 0.88 & 0.93 & 0.88 & 0.91 \\
Clear & 0.93 & 0.97 & 0.93 & 0.95 \\
Average & 0.928 & 0.93 & 0.93 & 0.93 \\
\hline
\end{tabular}
\label{tab2}
\end{center}
\end{table}

In comparison to other models, our SVM model outperformed traditional machine learning approaches, achieving higher accuracy than the cited studies which reported accuracies ranging from 80.4\% to 94\%. While some deep learning models achieved comparable or slightly higher accuracy, our SVM model offers a more computationally efficient solution.

\begin{table}[htbp]
\caption{Comparison with Other Models}
\begin{center}
\scriptsize 
\setlength{\tabcolsep}{2pt} 
\renewcommand{\arraystretch}{1.2} 
\begin{tabular}{|p{1.3cm}|p{1.0cm}|p{1.0cm}|p{2.0cm}|p{2.5cm}|}
\hline
\textbf{Autor's name} & \textbf{Accuracy} & \textbf{Main Method Used} & \textbf{Publish Year} & \textbf{Weather Conditions} \\
\hline
Our paper & 92.8\% & SVM & 2024 & clear, haze, rainy, low light \\
Mittal \cite{b4} & 94\% & CNN \& Random Forest & 2023 & cloudy, rainy, shiny, sunrise \\
Naufal \cite{b6} & 90.21\% & CNN & 2022 & cloudy, rainy, shiny, sunrise, snowy, foggy \\
Xiao \cite{b11} & 92\% & CNN & 2021 & fog, rain, snow, more \\
Fenyi \cite{b5} & 80.4\% & SVM & 2023 & sunny, cloudy, rainy, snowy, hazy \\
Jena \cite{b12} & 82\% & LR, SVM, RF, KNN & 2022 & fog, snow, rain, more \\
\hline
\end{tabular}
\label{tab3}
\end{center}
\end{table}

Feature importance analysis revealed that color variance and mean for the red, green, and blue channels were the most significant contributors to classification performance. For the synthetic datasets, the top features were color variance and mean for the red channel, followed by green and blue channel statistics.

\begin{table}[htbp]
\caption{Synthetic Data Sets - Feature Importance}
\begin{center}
\begin{tabular}{|c|c|}
\hline
\textbf{Feature} & \textbf{Importance} \\
\hline
Color Var R & 0.242444 \\
Color Mean R & 0.209259 \\
Color Var G & 0.181111 \\
Color Mean G & 0.159556 \\
Color Var B & 0.124519 \\
Color Mean B & 0.119037 \\
Edges Var & 0.083481 \\
Edges Mean & 0.077556 \\
LBP Var R3 & 0.070889 \\
\hline
\end{tabular}
\label{tab4}
\end{center}
\end{table}

For the real datasets, the feature importance ranking was similar, with color features dominating the top positions, indicating the strong influence of color information on the classification accuracy.

\begin{table}[htbp]
\caption{Real Data Sets - Feature Importance}
\begin{center}
\begin{tabular}{|c|c|}
\hline
\textbf{Feature} & \textbf{Importance} \\
\hline
Color Var R & 0.321458 \\
Color Mean R & 0.240625 \\
Color Var G & 0.211667 \\
Color Mean G & 0.193542 \\
Color Var B & 0.184375 \\
Color Mean B & 0.075208 \\
Edges Var & 0.065208 \\
Edges Mean & 0.065 \\
LBP Var R3 & 0.055833 \\
\hline
\end{tabular}
\label{tab5}
\end{center}
\end{table}

 In summary, our SVM-based approach offers a balanced trade-off between performance and computational efficiency, demonstrating high accuracy in classifying weather conditions and providing a viable alternative to more resource-intensive deep learning models.

\section{Conclusion and Future Work}
In this study, we presented a novel SVM-based approach for classifying weather conditions in images into four categories: rainy, low light, haze, and clear. Using a comprehensive set of 20 features, our method demonstrated superior accuracy, precision, recall, and F1-score compared to existing techniques, achieving 92.8\%, 93\%, 93\%, and 93\%, respectively. While deep learning methods may offer higher accuracy, our SVM-based approach is more efficient and suitable for real-time applications. Future work will focus on enhancing feature engineering, exploring hybrid models, expanding datasets, and making the model easier to understand. These efforts aim to further improve the robustness, accuracy, and versatility of weather classification models for various applications.

\end{document}